\documentclass[Afour,sageh,times]{sagej}

\usepackage{moreverb,url}
\usepackage[colorlinks,bookmarksopen,bookmarksnumbered,citecolor=red,urlcolor=red]{hyperref}
\usepackage{amssymb}
\usepackage{bm}
\usepackage[caption=false,font=normalsize,labelfont=sf,textfont=sf]{subfig}
\usepackage{comment}

\newcommand\BibTeX{{\rmfamily B\kern-.05em \textsc{i\kern-.025em b}\kern-.08em
T\kern-.1667em\lower.7ex\hbox{E}\kern-.125emX}}

\def\volumeyear{2016}

\begin{document}

\runninghead{Simultaneous estimation of contact position and tool shape}

\title{Simultaneous estimation of contact position and tool shape with high-dimensional parameters using force measurements and particle filtering}

\author{Kyo Kutsuzawa\affilnum{1} and Mitsuhiro Hayashibe\affilnum{1}}

\affiliation{\affilnum{1}Department of Robotics, Graduate School of Engineering, Tohoku University, Japan}

\corrauth{Kyo Kutsuzawa, Department of Robotics, Graduate School of Engineering, Tohoku University, 6-6-01 Aoba, Aramaki, Aoba-ku, Sendai, 980-8579, Japan.}

\email{kutsuzawa@ieee.org}

\begin{abstract}
Estimating the contact state between a grasped tool and the environment is essential for performing contact tasks such as assembly and object manipulation.
Force signals are valuable for estimating the contact state, as they can be utilized even when the contact location is obscured by the tool.
Previous studies proposed methods for estimating contact positions using force/torque signals; however, most methods require the geometry of the tool surface to be known.
Although several studies have proposed methods that do not require the tool shape, these methods require considerable time for estimation or are limited to tools with low-dimensional shape parameters.
Here, we propose a method for simultaneously estimating the contact position and tool shape, where the tool shape is represented by a grid, which is high-dimensional (more than 1000 dimensional).
The proposed method uses a particle filter in which each particle has individual tool shape parameters, thereby to avoid directly handling a high-dimensional parameter space.
The proposed method is evaluated through simulations and experiments using tools with curved shapes on a plane.
Consequently, the proposed method can estimate the shape of the tool simultaneously with the contact positions, making the contact-position estimation more accurate.
\end{abstract}


\maketitle

\begin{figure*}
    \centering
    \includegraphics{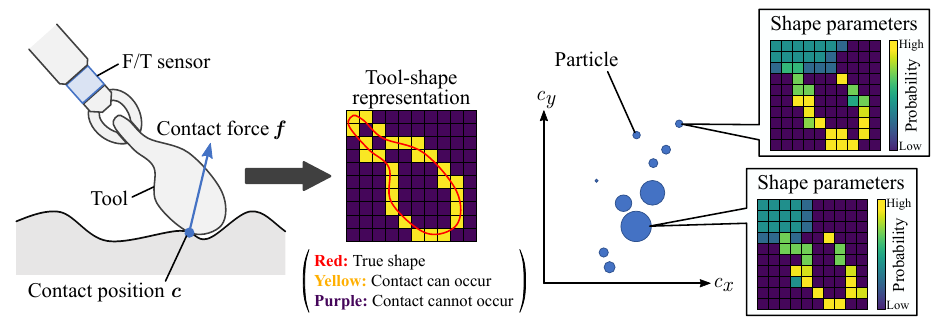}
    \caption{Method overview.
        \textbf{Left:} The proposed method assumes a single contact point on a rigid tool.
        \textbf{Middle:} The tool shape is represented as a grid map in which each cell encodes whether the tool surface exists in the cell.
        Here, yellow cells contain the tool surface, meaning a contact point can appear within the cell.
        \textbf{Right:} A particle filter estimates contact position using a set of weighted particles distributed in a space of contact position ($\bm{c} \triangleq [c_x, c_y]^{\top}$ in the figure).
        Each particle has its own grid map that represents shape parameters.
        Each time a new observation is made, the particle filter updates coordinates of the particles, their weights, and shape parameters. \label{fig:method-overview}}
\end{figure*}

\section{Introduction}

When robots manipulate objects or use them as tools, they often need to recognize the contact states between the grasped objects/tools and the environment.
For instance, when inserting a key into a keyhole, the robot must know the contact position and conditions of the key and lock.
In addition, when cutting the bone-in meat with a knife, the robot must detect where the knife contacts with the bone.
In such situations, robots need to detect contact states indirectly because tools usually have no sensors.
Moreover, because tools hide the contact location, robots must estimate the contact information from force signals instead of vision.

Conventional methods for contact-position estimation from force signals require the shape and position of the tools.
A technique for contact-position estimation \citep{Salisbury1984} often requires the shape of the tool surface to be known.
However, shape measurements using cameras generally exhibit large errors in the depth direction and are sensitive to occlusion, reflection, and transparency.
Although there are several methods for contact-position estimation without shape information \citep{Tsuji2017,Koike2017}, the estimation is slow, and these methods require that contact force constantly fluctuates during estimation.
As those drawbacks are unavoidable unless using shape information, it is beneficial to estimate tool shape for estimating the contact position.
Additionally, the shape of the object/tool is necessary to assemble tasks and plan a control strategy \citep{VonDrigalski2020}.

For tools made of transparent or reflective materials, it would be helpful to be able to estimate the tool shape from force signals instead of vision.
Recently, a method that simultaneously estimates the contact position and tool shape from force signals was proposed \citep{Kutsuzawa2020}.
This method gradually estimates the contact position and tool shape under uncertainty using an unscented particle filter (UPF) \citep{Merwe2000,VanDerMerwe2001}.
However, it requires the tool shape to be expressed using a small number of parameters.
It is practically impossible to apply that method to general shapes because the dimensionality of the tool-shape parameters becomes high, which requires an exponential number of particles for a reliable estimation.
There is another method that can detect the contact position while estimating the tool shape of voxels from force measurements \citep{Bimbo2022}, but this method requires the geometry of the environment being static.

Here, we propose a method to estimate the tool shape with a large number of parameters from force signals while simultaneously estimating the contact position, based on the Rao--Blackwellized particle filter (RBPF) \citep{Murphy2000} used in SLAM \citep{Murphy2000,Grisetti2005,Grisetti2007b}.
The conventional method \citep{Kutsuzawa2020} is affected by \emph{the curse of dimensionality} owing to the application of a particle filter to high-dimensional parameters.
By contrast, the proposed method avoids this issue by associating shape information with individual particles rather than scattering particles into the shape-parameter space.
Thus, the proposed method enables the tool shape to be expressed using a high-dimensional grid representation (voxels or pixels).
In addition, in contrast with \cite{Bimbo2022}, the proposed method does not need any assumption of environment geometry.
Therefore, the proposed method is available even for the contact object in the environment being deformable and movable, e.g., a human touching the tool.
In this study, we address the simple case of a curved object on a plane.
Although simple, this setup can demonstrate the effectiveness of the proposed method because the challenge of this study is the high-dimensional shape parameter.
The contributions of this study are listed as follows:
\begin{enumerate}
    \item This study proposes a force-signal-based estimation method for tool shapes represented by high-dimensional variables (grid representation) without any assumptions of environment geometry.
    \item We formulated a probabilistic model for contact-position and tool-shape estimation from force signals.
    \item This study also proposes a method for updating the tool-shape parameters represented by a grid using the estimated contact position and measurements.
\end{enumerate}

\section{Related Works}

Salisbury first proposed a method for estimating the contact position using six-axis force/torque signals \citep{Salisbury1984}.
This method is based on the equilibrium of the force and moment of force as follows:
\begin{align}
    \bm{F} &= \bm{f} \label{eqn:equilibrium of force},\\
    \bm{M} &= \bm{c} \times \bm{f} \label{eqn:equilibrium of moment},
\end{align}
where $\bm{F}$ and $\bm{M}$ indicate measured force and moment, and $\bm{c}$ and $\bm{f}$ indicate the contact position and force, respectively.
From these equilibrium equations, $\bm{c}$ can be determined as follows:
\begin{equation}
    \bm{c} = \frac{\bm{F} \times \bm{M}}{\|\bm{F}\|^2} + \alpha \bm{F} \label{eqn:line of action},
\end{equation}
where $\alpha$ is an unknown constant.
This formulation implies that the contact position cannot be uniquely determined from the force signals.
Instead, a line of contact-position candidates (\emph{a line of action}) is determined.
The method in \cite{Salisbury1984} estimates the unique contact position to be the intersection of the line and tool surface as follows:
\begin{equation}
    \bm{c} = \frac{\bm{F} \times \bm{M}}{\|\bm{F}\|^2} + \alpha \bm{F}, \quad \text{s.t. } g(\bm{c}) = 0 \label{eqn:contact position estimation with known shape}.
\end{equation}
A hypersurface with $g(\bm{p}) = 0$ represents the tool surface.
\cite{Bicchi1990} extended this method to soft-finger contacts.
As these methods do not require arrayed tactile sensors and enable force/torque sensors to be placed far from the object surface where contact occurs, they can be applied to the contact estimation of external objects and tools.
They can also be applied to the tactile sensing of robot bodies \citep{Galvez2001,Iwata2002,Tsuji2009,Kim2021}.
The accuracy of contact-position estimation can be improved by combining other information, such as tool velocity \citep{Muto1993a} and the derivative of force signals \citep{Kitamura2016}.
Some studies have extended these methods to multiple-point contact \citep{Featherstone1999,Kutsuzawa2015,Kutsuzawa2016,Manuelli2016}, and these techniques have been applied to robot control \citep{Featherstone1999,Park2008,Kutsuzawa2017a,Benallegue2018}.

Instantaneous measurements can only provide an estimate within the scope of (\ref{eqn:contact position estimation with known shape}), limiting the performance.
Therefore, as \cite{Salisbury1984} suggested, past measurements can be used to improve contact-position estimation.
\cite{Kurita2012} proposed a contact-position estimation method for non-convex shapes.
Extending the method proposed in \cite{Kurita2012}, \cite{Tsuji2017} proposed a contact-position estimation method without shape geometry.
\cite{Koike2017} implemented this method using a particle filter.

These methods have also been extended to estimate geometric information.
\cite{Mimura1994a} proposed a method for classifying contact states based on constrained degrees of freedom.
\cite{Karayiannidis2014} estimated the normal vector of a contact plane.
\cite{VonDrigalski2020} used a particle filter to identify the pose of a grasped tool with a known shape.
\cite{Liang2022} used a particle filter to identify the pose and shape of a grasped tool based on the force signals.
They also proposed a contact strategy to achieve fast estimation convergence.
Although they achieved high-precision estimation, their method required the tool to be a rigid column of known length and the contact plane to be known.
\cite{Tsujimura1989} proposed a method for detecting the external object geometry from force signals by touching using a probe, while it requires a probe shape, and assumes that the contact position is uniquely determined.
\cite{Bimbo2022} proposed a method that can reconstruct the geometry of tool and environment simultaneously with contact-position estimation.
However, their method requires the geometry of the environment being static.

Although particle filters have often been used for the localization of external objects \citep{Chhatpar2005,Petrovskaya2006,Platt2011}, Gadeyne et al. proposed a particle filter for simultaneous contact state and geometrical parameter estimation from force signals and hand-tip positions \citep{Gadeyne2005}.
They observed that the problem was similar to simultaneous localization and mapping (SLAM) \citep{Thrun1998} and employed a particle filter to address the \emph{chicken-and-egg} type estimation problem.
Their concept is similar to that of this study, whereas the geometrical parameters were up to 12-dimensional (the positions/orientations of the manipulated and environment objects).
The present study is based on \cite{Kutsuzawa2020} that proposed a method for the simultaneous estimation of contact position and tool shape.
The method proposed in \cite{Kutsuzawa2020} can perform estimations only from force signals; however, the tool shape parameter is only approximately two-dimensional.
Compared with these studies, the present study can handle high-dimensional (more than 1000 dimensions) shape parameters by employing techniques used in some SLAM methods \citep{Murphy2000,Grisetti2005,Grisetti2007b}.

\section{Method}

\subsection{Overview}

The proposed method simultaneously estimates the contact position and tool shape using a particle filter.
A particle filter is a Bayesian Monte Carlo technique that estimates variables based on the distribution of particles updated using Bayesian inference.
However, the proposed method does not directly handle the space of the tool-shape parameters;
instead, the tool-shape parameters are handled by associating them with each particle.
Therefore, the particles are distributed only in the contact-position space, making it easy to compute with a small number of particles, regardless of the dimensionality of the tool shape.
An overview of this process is shown in Figure~\ref{fig:method-overview}.

It should be noted that the proposed technique for handling high-dimensional shape parameters is based on the Rao--Blackwellized particle filter (RBPF) \citep{Murphy2000} used in SLAM \citep{Murphy2000,Grisetti2005,Grisetti2007b}.
In SLAM, the use of a grid map increases the number of dimensions of the variables to be estimated.
An RBPF is designed such that each particle, with an individual grid map computed from its own estimate, is scattered only in the localization space.
This significantly reduces the number of particles required.
The proposed method applies this technique to intrinsic contact sensing by designing a probabilistic model and a novel shape-parameter update method, as described below.

\subsection{Preliminary}

Let the contact position at the time step $k$ be $\bm{c}_k \in \mathbb{R}^3$.
Additionally, let an observation variable $\bm{y}_k \in \mathbb{R}^6$ comprise force $\bm{F}_k \in \mathbb{R}^3$ and the moment of force $\bm{M}_k \in \mathbb{R}^3$ measured by a force/torque sensor.
We assume that the following force equilibrium is maintained between the contact position and the measured values:
\begin{align}
    \bm{M}_k &= \bm{c}_k \times \bm{F}_k \label{eqn:equilibrium of moment with measured force}.
\end{align}
This formulation assumed a single contact point with no torque at the contact point.
Gravity and inertial forces were excluded from the measurements in advance.
Note that this method can also be applied to two-dimensional cases by omitting the $z$-axis components of position and force as well as the $x$- and $y$-axes components of the moment of force.

We define the object-shape parameter as $\bm{s}_k \in \mathbb{R}^{N_{\mathrm{s}}}$, where $N_{\mathrm{s}}$ indicates the number of object-shape parameters.
We let $\bm{s}_k$ be expressed by a grid map, where each cell takes $0$ to $1$;
the value is $1$ when the object surface is at the cell, and $0$ for no object surface.
This representation is high-dimensional;
when a region of $30\mathrm{~cm} \times 30\mathrm{~cm} \times 30\mathrm{~cm}$ is filled with $1\text{-cm}$ cell size, the number of cells is $N_{\mathrm{s}} = 30^3 = 27000$.
This is an extreme high dimensionality compared with conventional studies \citep{Kutsuzawa2020,Gadeyne2005}, in which the dimensions of the shape parameters were two or twelve.

For the sake of simplicity, we combined the variables to be estimated at time $k$, i.e. $\bm{c}_k$ and $\bm{s}_k$, into a state variable $\bm{x}_k$ as follows:
\begin{equation}
    \bm{x}_k \triangleq (\bm{c}_k, \bm{s}_k).
\end{equation}
Additionally, we denote $\bm{x}_{0:k}$ as a sequence of variables $\bm{x}$ from the time step to $0$ to $k$ as follows:
\begin{equation}
    \bm{x}_{0:k} \triangleq (\bm{x}_0, \bm{x}_1, \ldots, \bm{x}_k).
\end{equation}
This notation is also used for other time-series variables in the same way, e.g. $\bm{y}_{0:k} \triangleq (\bm{y}_0, \bm{y}_1, \ldots, \bm{y}_k)$.

\subsection{Probabilistic model}

The generative model of the system is formulated in a recursive way as follows:
\begin{align}
    &p(\bm{x}_{0:k}, \bm{y}_{0:k}) \nonumber\\
    &= p(\bm{y}_k| \bm{x}_k) p(\bm{x}_k| \bm{x}_{k-1}) p(\bm{x}_{0:k-1}, \bm{y}_{0:k-1}),
\end{align}
where
\begin{align}
    p(\bm{y}_k| \bm{x}_k) &= p(\bm{M}_k| \bm{c}_k, \bm{F}_k) p(\bm{F}_k), \nonumber\\
    p(\bm{x}_k| \bm{x}_{k-1}) &= p(\bm{c}_k| \bm{c}_{k-1}, \bm{s}_k) p(\bm{s}_k| \bm{s}_{k-1}).
\end{align}

As we would like to know $\bm{x}_{0:k}$, the goal is to estimate the following expectation value:
\begin{align}
    &\mathbb{E}_{\bm{x}_{0:k} \sim p(\bm{x}_{0:k} | \bm{y}_{0:k})}[\phi(\bm{x}_{0:k})] \nonumber\\
    &= \int \phi(\bm{x}_{0:k}) p(\bm{x}_{0:k} | \bm{y}_{0:k}) \mathrm{d}\bm{x}_{0:k},
\end{align}
where $\phi$ denotes an arbitrary function.
For instance, by defining as $\phi(\bm{x}_{0:k}) = \bm{c}_k$ and $\phi(\bm{x}_{0:k}) = \bm{s}_k$, we can obtain the expected contact position and tool shape at time $k$, respectively.

\subsection{Particle filtering algorithm}

\subsubsection{Naive particle filtering}

In the particle filters, the expectation of $p(\bm{x}_{0:k} | \bm{y}_{0:k})$ is replaced with the expectation of an arbitrary distribution $q(\bm{x}_{0:k} | \bm{y}_{0:k})$, which is referred to as \emph{the proposal distribution}, as follows:
\begin{align}
    &\mathbb{E}_{\bm{x}_{0:k} \sim p(\bm{x}_{0:k} | \bm{y}_{0:k})}[\phi(\bm{x}_{0:k})] \nonumber\\
    &= \int \phi(\bm{x}_{0:k}) p(\bm{x}_{0:k} | \bm{y}_{0:k}) \mathrm{d}\bm{x}_{0:k} \nonumber\\
    &= \frac{\int \phi(\bm{x}_{0:k}) w_k q(\bm{x}_{0:k}| \bm{y}_{0:k}) \mathrm{d}\bm{x}_{0:k}}{\int w_k q(\bm{x}_{0:k}| \bm{y}_{0:k}) \mathrm{d}\bm{x}_{0:k}} \nonumber\\
    &= \frac{\mathbb{E}_{\bm{x}_{0:k} \sim q(\bm{x}_{0:k}| \bm{y}_{0:k})} [\phi(\bm{x}_{0:k}) w_k]}{\mathbb{E}_{\bm{x}_{0:k} \sim q(\bm{x}_{0:k}| \bm{y}_{0:k})} [w_k]} \label{eqn:naive filtering},
\end{align}
where $w_k$ denotes a weight variable defined as follows:
\begin{equation}
    w_k \triangleq \frac{p(\bm{y}_{0:k}| \bm{x}_{0:k}) p(\bm{x}_{0:k})}{q(\bm{x}_{0:k}| \bm{y}_{0:k})}.
\end{equation}
Moreover, by assuming
\begin{align}
    q(\bm{x}_{0:k} | y_{0:k})
    &= q(\bm{x}_k | \bm{x}_{0:k-1}, \bm{y}_{0:k}) q(\bm{x}_{0:k-1} | \bm{y}_{0:k}) \nonumber\\
    &= q(\bm{x}_k | \bm{x}_{0:k-1}, \bm{y}_{0:k}) q(\bm{x}_{0:k-1} | \bm{y}_{0:k-1}),
\end{align}
we can rewrite the weight variables as follows:
\begin{equation}
    w_k = \frac{p(\bm{y}_k | \bm{x}_k) p(\bm{x}_k | \bm{x}_{k-1})}{q(\bm{x}_k | \bm{x}_{0:k-1}, \bm{y}_{0:k})} w_{k-1} \label{eqn:naive weight}.
\end{equation}
Finally, the expectation of (\ref{eqn:naive filtering}) is approximated by using the weighted mean of samples from the proposal distribution as follows:
\begin{equation}
    \mathbb{E}_{\bm{x}_{0:k} \sim p(\bm{x}_{0:k} | \bm{y}_{0:k})}[\phi(\bm{x}_{0:k})] \approx \frac{\sum_{i=0}^{N-1} \phi(\bm{x}^{(i)}_{0:k}) w^{(i)}_k}{\sum_{i=0}^{N-1} w^{(i)}_k} \label{eqn:particle approximation},
\end{equation}
where
\begin{align}
    w^{(i)}_k &= \frac{p(\bm{y}_k | \bm{x}^{(i)}_k) p(\bm{x}^{(i)}_k | \bm{x}^{(i)}_{k-1})}{q(\bm{x}^{(i)}_k | \bm{x}^{(i)}_{0:k-1}, \bm{y}_{0:k})} w^{(i)}_{k-1} \label{eqn:naive weight approximation},\\
    \bm{x}^{(i)}_k &\sim q(\bm{x}^{(i)}_k | \bm{x}^{(i)}_{0:k-1}, \bm{y}_{0:k}) \label{eqn:naive particle sampling}.
\end{align}

\subsubsection{Proposed particle filtering}

A naive particle filter is based on (\ref{eqn:particle approximation})--(\ref{eqn:naive particle sampling}) to compute the expectation value using the average of the samples (particles).
However, this is practically impossible owing to the high dimensionality of shape parameters.
There are two main difficulties with this method:
1) Many particles are required for a valid approximation in (\ref{eqn:particle approximation}), which is difficult to handle exponentially as the number of dimensions increases owing to the curse of dimensionality;
2) Updating the proposal distribution, which must handle $\dim \bm{x} \times \dim \bm{x}$ covariance matrices, is computationally heavy as $\dim \bm{x}$ increases.
Therefore, the implementation of particle filtering must be modified to avoid sampling from shape-parameter space.

We first modify the proposal distribution as follows:
\begin{align}
    &q(\bm{x}_k | \bm{x}_{0:k-1}, \bm{y}_{0:k}) \nonumber\\
    &\triangleq q(\bm{c}_k, \bm{s}_k | \bm{c}_{0:k-1}, \bm{s}_{0:k-1}, \bm{M}_{0:k}, \bm{F}_{0:k}) \nonumber\\
    &= q(\bm{c}_k | \bm{c}_{0:k-1}, \bm{M}_{0:k}, \bm{F}_{0:k}) q(\bm{s}_k | \bm{s}_{k-1}, \bm{c}_{k-1}, \bm{F}_{k-1}).
\end{align}
Although this is an arbitrary decomposition, we can separate the probabilistic distribution of $\bm{c}_k$ and that of  $\bm{s}_k$.

We also define the following assumption:
the shape parameter $\bm{s}_k$ can be uniquely updated from the contact position $\bm{c}_{k-1}$ and the measured force $\bm{F}_{k-1}$.
This is reasonable because the contact position is always on the shape surface and the contact force vector is usually directed inward toward the object (i.e., only the pushing force can be applied).
This assumption is formulated as follows:
\begin{align}
    &q(\bm{s}_k | \bm{s}_{k-1}, \bm{c}_{k-1}, \bm{F}_{k-1}) \nonumber\\
    &=
    \begin{cases}
        1 & \text{if } \bm{s}_k = \psi(\bm{s}_{k-1}, \bm{c}_{k-1}, \bm{F}_{k-1}), \\
        0 & \text{otherwise},
    \end{cases}
\end{align}
where $\psi$ denotes an update function of the shape parameter which is defined in detail later.

By employing this assumption, we can obtain particles $\{\bm{x}^{(i)}_k\}_{i = 1, \ldots, N}$ as follows:
\begin{align}
    \bm{c}^{(i)}_k &\sim q(\bm{c}^{(i)}_k | \bm{c}^{(i)}_{0:k-1}, \bm{M}_{0:k}, \bm{F}_{0:k}) \label{eqn:proposed sampling of c},\\
    \bm{s}^{(i)}_k &= \psi(\bm{s}_{k-1}, \bm{c}_{k-1}, \bm{F}_{k-1}) \label{eqn:proposed sampling of s}.
\end{align}
Here, superscript $(i)$ indicates the index of the particles sampled from the proposal distribution.
We use the UPF \citep{Merwe2000,VanDerMerwe2001}, which employs an unscented Kalman filter (UKF) \citep{Julier1997,Wan2000}, for implementing the proposal distribution, $q(\bm{x}^{(i)}_k | \bm{x}^{(i)}_{0:k-1}, \bm{y}_{0:k})$.

Then, we can compute (\ref{eqn:naive filtering}) using (\ref{eqn:particle approximation}), where
\begin{equation}
    w^{(i)}_k = \frac{p(\bm{M}_k, \bm{F}_k | \bm{c}^{(i)}_k) p(\bm{c}^{(i)}_k | \bm{c}^{(i)}_{k-1}, \bm{s}^{(i)}_k)}{q(\bm{c}^{(i)}_k | \bm{c}^{(i)}_{0:k-1}, \bm{M}_{0:k}, \bm{F}_{0:k})} w^{(i)}_{k-1}.
\end{equation}
We modeled probabilistic distributions as follows:
\begin{align}
    p(\bm{M}_k, \bm{F}_k | \bm{c}^{(i)}_k) &= p(\bm{M}_k | \bm{F}_k, \bm{c}^{(i)}_k) p(\bm{F}_k),\\
    p(\bm{M}_k | \bm{F}_k, \bm{c}^{(i)}_k) &= \mathcal{N}(\bm{M}_k ; \bm{c}^{(i)}_k \times \bm{F}_k, \bm{\Sigma}^{\bm{M}}),\\
    p(\bm{F}_k) &= \mathrm{const},\\
    p(\bm{c}^{(i)}_k | \bm{c}^{(i)}_{k-1}, \bm{s}^{(i)}_k) &= p(\bm{c}^{(i)}_k | \bm{c}^{(i)}_{k-1}) p(\bm{c}^{(i)}_k | \bm{s}^{(i)}_k),\\
    p(\bm{c}^{(i)}_k | \bm{c}^{(i)}_{k-1}) &= \mathcal{N}(\bm{c}^{(i)}_k ; \bm{c}^{(i)}_{k-1}, \bm{\Sigma}^{\bm{c}}),\\
    p(\bm{c}^{(i)}_k | \bm{s}^{(i)}_k) &= \frac{\exp\left[ \mathrm{Grid}_{\bm{s}^{(i)}_k}(\bm{c}^{(i)}_k) \right]}{\int \exp\left[ \mathrm{Grid}_{\bm{s}^{(i)}_k}(\bm{c})  \right] \mathrm{d} \bm{c}}.
\end{align}
Here, $\mathcal{N}(\bm{x} ; \bm{\mu}, \bm{\Sigma})$ indicates the probability distribution of $\bm{x}$ following a Gaussian distribution with its mean $\bm{\mu}$ and its variance $\bm{\Sigma}$.
$\bm{\Sigma}^{\bm{M}} = \sigma_{\bm{M}}^2 \bm{I}$ and $\bm{\Sigma}^{\bm{c}} = \sigma_{\bm{c}}^2 \bm{I}$ indicate the variance of the moment and contact position, respectively.
Additionally, $\mathrm{Grid}_{\bm{s}}(\bm{x})$ indicates the value of the cell of the grid $\bm{s}$ that contains position $\bm{x}$.

Finally, we compute the particle filtering (right side of (\ref{eqn:particle approximation})) without sampling from the high-dimensional shape-parameter space.
Instead, we sampled $\bm{c}^{(i)}_k$ from the proposal distribution and computed $\bm{s}^{(i)}_k$ for each particle.

In summary, the main difference between the naive and proposed filtering algorithms is the sampling strategy, as can be seen by comparing (\ref{eqn:naive particle sampling}) and (\ref{eqn:proposed sampling of c})--(\ref{eqn:proposed sampling of s}).
In the naive particle filter, the proposal distribution $q$ needs to handle a high-dimensional space, but in the proposed particle filter, by assuming a deterministic shape update rule, the proposal distribution only needs to handle a low-dimensional space instead.

\subsection{Shape estimation}

\begin{figure}
    \centering
    \subfloat[Shape-estimation criteria.]{
        \includegraphics[scale=0.95]{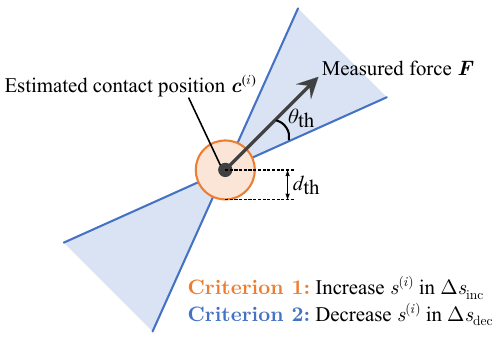}
    }\\
    \subfloat[Grid-map increments.]{
        \includegraphics[scale=0.95]{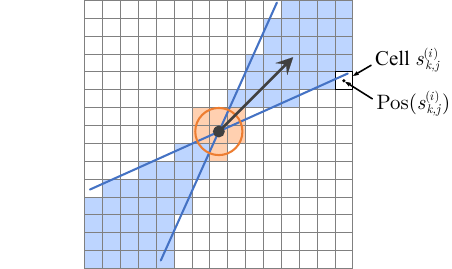}
    }
    \caption{Overview of tool-shape computation.
        \textbf{(a)} The criterion 1 (orange) increases shape parameters near the estimated contact position. The criterion 2 (blue) decreases shape parameters near the line of action expressed in (\ref{eqn:line of action}) to accelerate shape estimation.
        \textbf{(b)} The value of each cell is updated when the center of the cell is within the area of the criteria 1 or 2. The grid map corresponds to those illustrated in Figure~\ref{fig:method-overview}.
        \label{fig:shape computation}}
\end{figure}

To estimate the shape information effectively, the design of the function $\psi$ is important.
We design $\psi$ using the following two criteria.
\begin{enumerate}
    \item Tool surface exists around near the contact point $\bm{c}_{k-1}$.
    \item Tool surface does not exist where it is far from $\bm{c}_{k-1}$ along $\bm{F}_{k-1}$.
\end{enumerate}
The first criterion was based on the fact that the contact position was always on a tool surface.
The second criterion assists the estimation by reducing the existence of the tool shape at the point where no contact occurs.
Because the contact provides only local information, the second criterion, which changes a wide range of shape information, can accelerate the estimation.
An overview of these criteria is shown in Figure~\ref{fig:shape computation}.

The first criterion is implemented as follows:
\begin{align}
    s^{(i)}_{k, j} =& s^{(i)}_{k-1, j} + \Delta s_{\mathrm{inc}}, \nonumber\\
    &\text{if } \|\mathrm{Pos}(s^{(i)}_{k, j}) - \bm{c}^{(i)}_{k-1}\| < d_{\mathrm{th}} \label{eqn:shape estimation criterion 1}.
\end{align}
Here, $j$ indicates the cell index.
$\Delta s_{\mathrm{inc}}$ and $d_{\mathrm{th}}$ indicate the increment of the shape parameter and the threshold of distance, respectively.
$\mathrm{Pos}(x)$ indicates the center position of cell $x$.
This operation increases the values of the cell $s^{(i)}_{k, j}$ whose distance from the contact point $\bm{c}^{(i)}_{k-1}$ is less than $d_{\mathrm{th}}$.

The second criterion is implemented as follows:
\begin{align}
    s^{(i)}_{k, j} =& s^{(i)}_{k-1, j} - \Delta s_{\mathrm{dec}}, \nonumber\\
    &\text{if } \mathrm{Pos}(s^{(i)}_{k, j}) \in \mathrm{cone}(\bm{c}^{(i)}_{k-1}, \bm{F}_{k-1}, \theta_{\mathrm{th}}) \nonumber\\
    &\text{and } \|\mathrm{Pos}(s^{(i)}_{k, j}) - \bm{c}^{(i)}_{k-1}\| \geq d_{\mathrm{th}} \label{eqn:shape estimation criterion 2}.
\end{align}
Here, $\mathrm{cone}(\bm{a}, \bm{v}, \theta)$ indicates a double-cone with apex $\bm{a}$, axis along $\bm{v}$, and angle $\theta$.
$\Delta s_{\mathrm{dec}}$ and $\theta_{\mathrm{th}}$ indicate the decrement of the shape parameter and the threshold of the angle, respectively.
This operation decreases the values of cell $s^{(i)}_{k, j}$ that are near the line along $\bm{F}_{k-1}$ through $\bm{c}^{(i)}_{k-1}$.

\section{Simulation}

\subsection{Setup} \label{sec:simulation setup}

\begin{figure}
    \centering
    \includegraphics{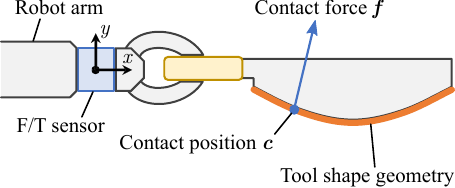}
    \caption{Setup for simulation. The simulation assumes that a robot arm with a force/torque sensor stably grasps a tool. The tool will contact with the environment at a single point within the orange area. \label{fig:tool for simulation}}
\end{figure}

\begin{table}
    \caption{Hyper-parameters of the proposed method in the simulation \label{tab:hyper-parameters}}
    \small\sf\centering
    \begin{tabular}{lcl}
        \toprule
        Item                       & Symbol                    & Value \\
        \midrule
        Number of particles        & $N$                       & $300$ \\
        Size of grid cells         & -                         & $0.5 \mathrm{~cm}$ \\
        Number of cells            & $N_{\mathrm{s}}$          & $80 \times 80$ \\
        Threshold of resampling    & $N_{\mathrm{th}}$         & $0.432$ \\
        Std. of contact position   & $\sigma_{\bm{c}}$         & $5.25 \times 10^{-4} \mathrm{~cm}$ \\
        Std. of measured moment    & $\sigma_{\bm{M}}$         & $3.79 \times 10^{-4} \mathrm{~Nm}$ \\
        Threshold of distance      & $d_{\mathrm{th}}$         & $0.939 \mathrm{~cm}$ \\
        Threshold of angle         & $\theta_{\mathrm{th}}$    & $0.108 \mathrm{~rad}$ \\
        Increment of shape params. & $\Delta s_{\mathrm{inc}}$ & $0.0347$ \\
        Decrement of shape params. & $\Delta s_{\mathrm{dec}}$ & $0.0216$ \\
        \bottomrule
    \end{tabular}
\end{table}

\begin{table}
    \caption{Hyper-parameters of the baseline method in the simulation, where the symbols follow \cite{Tsuji2017} \label{tab:hyper-parameters baseline}}
    \small\sf\centering
    \begin{tabular}{lcl}
        \toprule
        Item                  & Symbol   & Value \\
        \midrule
        Forgetting factor     & $\rho$   & $0.992$ \\
        Regularization factor & $\alpha$ & $860$ \\
        \bottomrule
    \end{tabular}
\end{table}

\begin{table}
    \caption{Hyper-parameters of the naive method \label{tab:hyper-parameters naive}}
    \small\sf\centering
    \begin{tabular}{lcl}
        \toprule
        Item                     & Symbol            & Value \\
        \midrule
        Number of particles      & $N$               & $300$ \\
        Size of grid cells       & -                 & $8.0 \mathrm{~cm}$ \\
        Number of cells          & $N_{\mathrm{s}}$  & $5 \times 5$ \\
        Threshold of resampling  & $N_{\mathrm{th}}$ & $0.432$ \\
        Std. of contact position & $\sigma_{\bm{c}}$ & $0.0114 \mathrm{~cm}$ \\
        Std. of shape parameters & $\sigma_{\bm{s}}$ & $4.37 \times 10^{-3}$ \\
        Std. of measured moment  & $\sigma_{\bm{M}}$ & $1.69 \times 10^{-5} \mathrm{~Nm}$ \\
        \bottomrule
    \end{tabular}
\end{table}

We evaluated the proposed method through a simulation as illustrated in Figure~\ref{fig:tool for simulation}.
It simulated a robotic arm with an F/T sensor gripping a knife-shaped tool in a plane.
The simulator was implemented by Python, and simulates the following contents.
The contact surface has a shape defined as follows:
\begin{equation}
    \big\{ (x, y) \big| 0.1 \mathrm{~m} \leq x \leq 0.3 \mathrm{~m}, y = h(x) \big\},
\end{equation}
where $h: \mathbb{R} \rightarrow \mathbb{R}$ is a function that specifies the tool shape.
The estimation was performed based on the coordinate system of the F/T sensor.
We assume that gravitational and inertial forces are eliminated.
Thus, the estimation can be performed without considering robot motion.

\begin{figure*}
    \centering
    \includegraphics[scale=0.95]{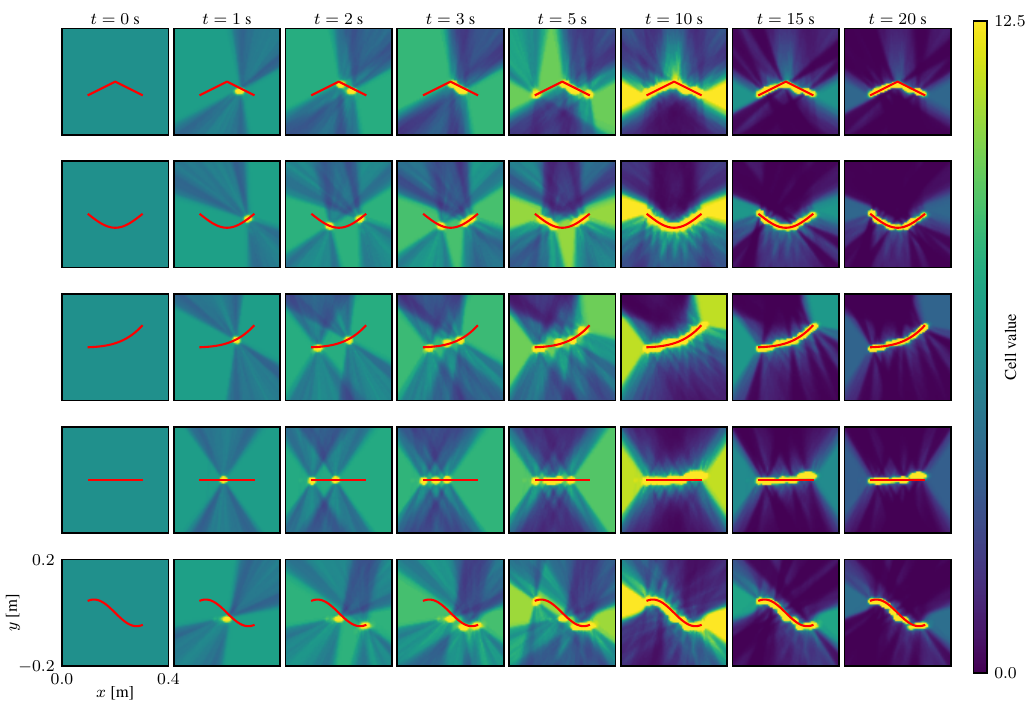}
    \caption{Progress of shape parameter estimation in the proposed method.
        The color maps represent the estimated tool shape, where yellow indicates a high degree of certainty that the tool surface is there (i.e., contact can occur).
        Red curves indicate the true object shape.
        Here, the cell value corresponds to $s_k$, and the value of 12.5 in the color bar was arbitrarily chosen for the sake of visibility in the color map. \label{fig:shape parameter estimation progress}}
\end{figure*}

\begin{figure*}
    \centering
    \includegraphics[scale=0.95]{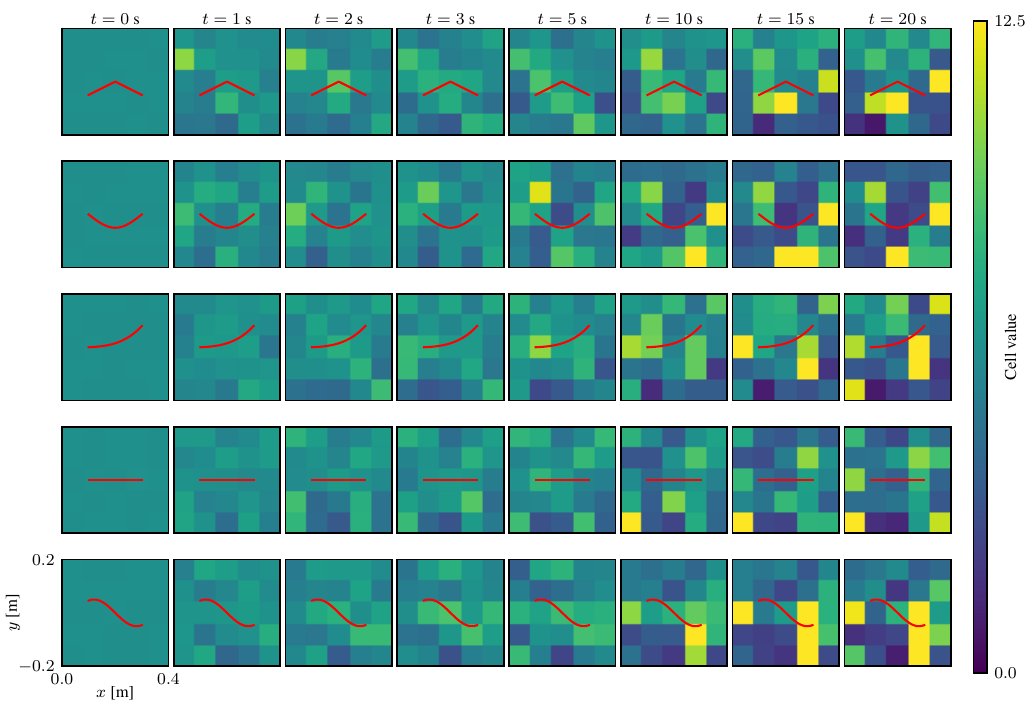}
    \caption{Progress of shape parameter estimation in the naive method.
        The color maps represent the estimated tool shape, where yellow indicates a high degree of certainty that the tool surface is there (i.e., contact can occur).
        Red curves indicate the true object shape. \label{fig:shape parameter estimation progress naive}}
\end{figure*}

\begin{figure}
    \centering
    \includegraphics{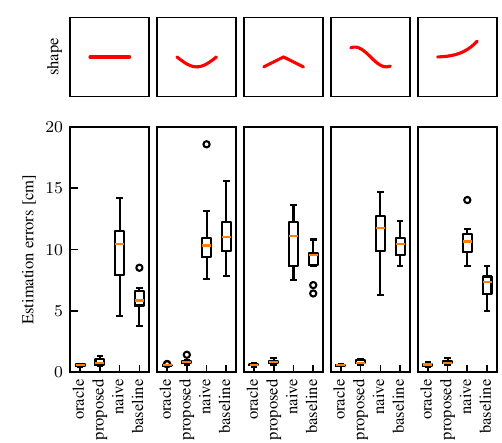}
    \caption{Estimation errors of the contact position after 10~s, where no fluctuation in contact force.
        Quartiles of 10 trials for each method and shape are shown. \label{fig:position errors results}}
\end{figure}

\begin{figure}
    \centering
    \subfloat[Proposed method]{
        \includegraphics[scale=0.95]{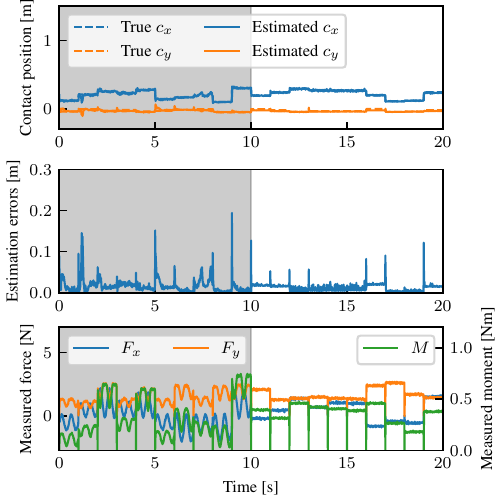}
        \label{fig:estimated position results proposed}
    }\\
    \subfloat[Baseline method]{
        \includegraphics[scale=0.95]{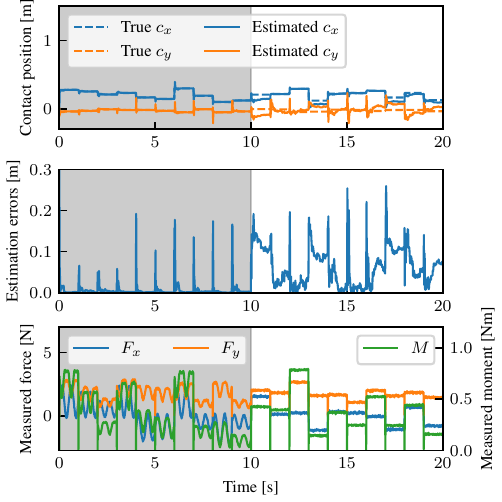}
        \label{fig:estimated position results baseline}
    }
    \caption{Estimated contact positions during simulation.
    The contact force intentionally fluctuated before 10~s (gray area), and then was maintained constant.
    \label{fig:estimated position results}}
\end{figure}

\begin{figure}
    \centering
    \subfloat[Estimated shape parameters.
    The red curves indicate the true shapes of the tools.]{
        \includegraphics[scale=0.95]{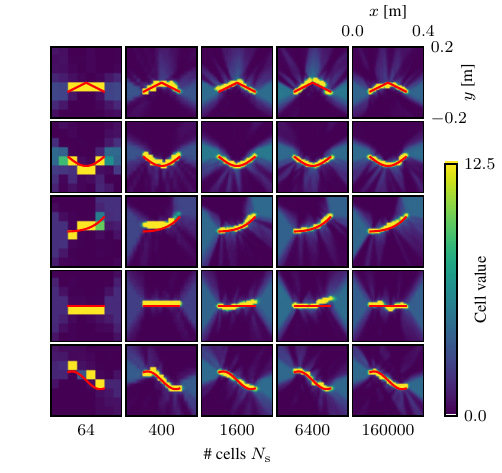}
        \label{fig:estimated shape parameters}
    }\\
    \subfloat[Position-estimation errors.
    The mean (solid curve) and standard deviation (filled area) of 10 trials are shown.]{
        \includegraphics[scale=0.95]{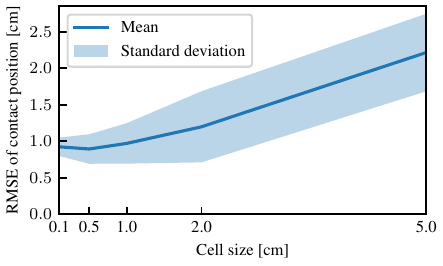}
        \label{fig:position estimation errors}
    }
    \caption{Estimation results with varying resolutions, $N_{\mathrm{s}}$. \label{fig:results with varying resolutions}}
\end{figure}

\begin{figure}
    \centering
    \includegraphics{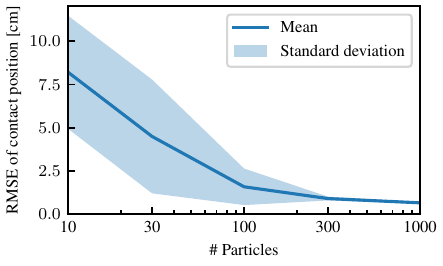}
    \caption{Estimation errors of the contact position with different numbers of particles, $N$.
        The mean (solid curve) and standard deviation (filled area) of 10 trials are shown. \label{fig:position errors results with varying particles}}
\end{figure}

\begin{figure}
    \centering
    \includegraphics{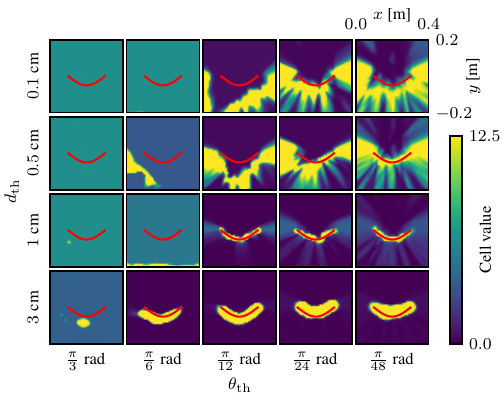}
    \caption{Estimated shape parameters with varying $d_{\mathrm{th}}$ and $\theta_{\mathrm{th}}$.
        The red curves indicate the true shapes of the tools. \label{fig:estimated shape parameters with varying params}}
\end{figure}

\begin{figure}
    \centering
    \includegraphics{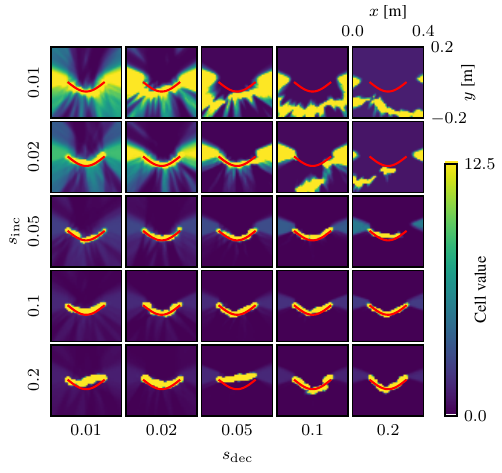}
    \caption{Estimated shape parameters with varying $\Delta s_1$ and $\Delta s_2$.
        The mean values of all particles are shown.
        The red curves indicate the true shapes of the tools. \label{fig:estimated shape parameters with varying delta s}}
\end{figure}

\begin{figure}
    \centering
    \includegraphics{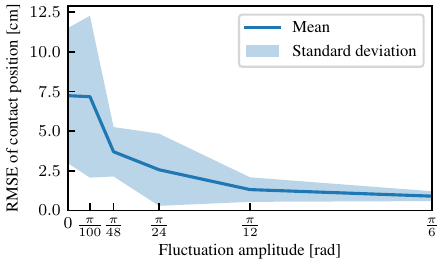}
    \caption{Estimation errors of the contact position with varying force-fluctuation amplitudes.
    The mean (solid curve) and standard deviation (filled area) of 10 trials are shown. \label{fig:position errors results with varying fluctuations}}
\end{figure}

Various external forces are applied at random positions on the tool surface.
The contact force fluctuated in the first 10~s, but stopped fluctuating thereafter.
During the task, the contact position $\bm{c} = (c_x, c_y)$ changes every 1~s randomly.
In the first 10~s, the contact force $\bm{f} = (f_x, f_y)$ varies as follows:
\begin{align}
    f_x &= A \sin \theta_f(t),\\
    f_y &= A \cos \theta_f(t),\\
    \theta_f(t) &= \theta_0 \sin (4 \pi t) + \theta_{\perp},
\end{align}
where the amplitude $A$ changes every 1~s as follows:
\begin{equation}
    A \sim U(1.0 \mathrm{~N}, 3.0 \mathrm{~N}).
\end{equation}
$\theta_{\perp}$ indicates the angle normal to the tool surface at the contact position $\bm{c}_k$, $t$ indicates the time, and $\theta_0$ indicates the amplitude of contact force fluctuation.
We set $\theta_0 = \frac{\pi}{6}$ unless otherwise specified.
After 10~s, the contact force was maintained constant for 1~s and changed every 1~s as follows:
\begin{align}
    f_x &= A \sin \theta_f,\\
    f_y &= A \cos \theta_f,\\
    \theta_f &\sim U\left(-\frac{\pi}{6}, \frac{\pi}{6}\right) + \theta_{\perp}.
\end{align}
We set the contact force vector within a certain angle of the tool surface normal, because stable contact requires the contact force vector to be within the friction cone.

The following three methods were used to compare the performance of contact-position and/or tool-shape estimation with the proposed method.
\begin{enumerate}
    \item Baseline: contact-position estimation method that does not require shape information \citep{Tsuji2017}.
    \item Naive: simultaneous contact-position and shape estimation with a naive particle filter \citep{Kutsuzawa2020}.
    \item Oracle: contact-position estimation method where the actual tool shape is provided \citep{Salisbury1984}.
\end{enumerate}
The hyper-parameters are listed in Tables~\ref{tab:hyper-parameters}--\ref{tab:hyper-parameters naive}, that were selected by using Optuna \citep{optuna}, a hyper-parameter optimization framework.
Note that the Oracle method does not have hyper-parameters; it estimates contact positions based on (\ref{eqn:contact position estimation with known shape}).

The shape estimation errors were computed by selecting the highest-value cells along the $y$-axis and calculating the distance between the center of the cells and the $y$-axis position of the true shape surface.
\begin{equation}
    e_{\mathrm{shape}} \triangleq \frac{1}{M} \sum_{\mu=1}^M \left| h(x_{\mu}) - \mathrm{Pos}\left( \operatorname{arg~max}_{\nu} s_{(\nu, \mu)} \right) \right|.
\end{equation}
Here, we did not use other common metrics for measuring differences between two shapes, such as Hausdorff distance.
Hausdorff distance, for example, calculates the maximum distance from a point on one shape to the closest point on the other shape, that is sensitive to outliers.
Also, common metrics generally require binarization in the continuous-valued shape parameters with an arbitrary threshold before they are applied.
This metric, on the other hand, does not need any threshold.

\subsection{Results}

\subsection{Performance of the proposed method}

We first show the performance of the proposed method with tuned hyper-parameters.
Also, we compared the proposed method with the other methods.

Figure~\ref{fig:shape parameter estimation progress} shows the progress of shape estimation in the proposed method in five types of shape.
Also, Figure~\ref{fig:shape parameter estimation progress naive} shows the shape-estimation progress in the naive particle filter.
In the proposed method, the tool-shape parameters gradually converged to near the true value.
On the other hand, the naive particle filter failed to estimate the tool shape although the resolution was quite low.
It should be noted that we could not run the naive particle filtering with $N = 300$ and $N_{\mathrm{s}} = 6400$ in our computational environment since it required approximately 366~GB of computer memory owing to the high-dimensionality of the state space.
Table~\ref{tab:shape-estimation errors} compares shape-estimation errors in the proposed and naive methods in each shape.
The proposed method resulted in approximately 0.5~cm of errors, whereas the naive method resulted in errors larger than 8~cm.

We then compared the estimation errors of contact positions between the proposed, naive, baseline, and oracle methods.
Figure~\ref{fig:position errors results} shows the estimation errors for the 10 trials.
The oracle method naturally achieved the highest performance, as it was provided with the true tool surface.
We also observed that the proposed method outperformed the other methods.

Figure~\ref{fig:estimated position results} presents the estimation results of contact positions in the proposed and baseline method.
In the baseline method, the estimated contact position was unstable and did not converge to the true values after 10~s, when the contact force stopped oscillating.
By contrast, the proposed method maintained approximately the same accuracy even after the contact force stopped oscillating.

\begin{table}
    \caption{Shape-estimation errors (means and standard deviations of 10 trials) \label{tab:shape-estimation errors}}
    \small\sf\centering
    \begin{tabular}{lcc}
        \toprule
        Shape    & \multicolumn{2}{c}{Errors [cm]} \\
                 & Proposed           & Naive \\
        \midrule
        Straight & $ 0.540 \pm 0.183$ & $ 8.800 \pm 2.680$ \\
        Arch     & $ 0.608 \pm 0.172$ & $ 8.600 \pm 3.191$ \\
        Angular  & $ 0.568 \pm 0.167$ & $ 9.600 \pm 3.809$ \\
        Wavy     & $ 0.694 \pm 0.177$ & $ 9.867 \pm 3.721$ \\
        Knife    & $ 0.585 \pm 0.132$ & $10.035 \pm 3.611$ \\
        \bottomrule
    \end{tabular}
\end{table}

\subsection{Hyper-parameter evaluation}

Up to this point, we have examined the performance of the proposed method in the optimized hyper-parameters listed in Table~\ref{tab:hyper-parameters}.
Hereinafter, we investigate how the proposed method behaves as the hyper-parameters are varied.

We firstly evaluated varied resolution of shape parameters, $N_{\mathrm{s}}$.
Figure~\ref{fig:results with varying resolutions}\subref{fig:estimated shape parameters} shows the estimated shape parameters at 20~s.
The proposed method could estimate even $N_{\mathrm{s}} = 400 \times 400 = 160000$ dimensions of the shape parameter using 300 particles.
In most cases, the shape surfaces were properly estimated regardless of the resolution.
This demonstrates the ability of the proposed method to handle shape parameters with significantly higher dimensions than the number of particles.
Figure~\ref{fig:results with varying resolutions}\subref{fig:position estimation errors} also shows the shape estimation errors with varying resolutions (cell sizes).
The estimation errors increased as the cell size increased (i.e., the resolution becomes coarse) and finally converged to about a half of the cell size.

We then evaluated with different numbers of particles, $N$.
Figure~\ref{fig:position errors results with varying particles} shows the position-estimation errors for varying numbers of particles.
Estimation errors decreased as the numbers of particles increase and almost converged in 300 particles.

We also evaluated hyper-parameters specific to the shape estimation.
Figure~\ref{fig:estimated shape parameters with varying params} shows estimated shape parameters in the wavy-shaped tool with varying $d_{\mathrm{th}}$ and $\theta_{\mathrm{th}}$.
Figure~\ref{fig:estimated shape parameters with varying delta s} shows estimated shape parameters in the wavy-shaped tool with varying $\Delta s_1$ and $\Delta s_2$.
For those parameters, the performance was degraded when the values were too large or too small.

The proposed method uses fluctuation of the contact force as a clue when the shape is totally unknown.
Therefore, although not a parameter of the proposed method itself, we also evaluated with varying amplitude of force fluctuations, $\theta_0$.
Figure~\ref{fig:position errors results with varying fluctuations} shows the estimation errors of contact position for varying amplitude of contact force fluctuation, $\theta_0$.
We can observe that the estimation error increases as the fluctuation decreases.

\section{Experiment}

\subsection{Setup}

\begin{figure}
    \centering
    \includegraphics{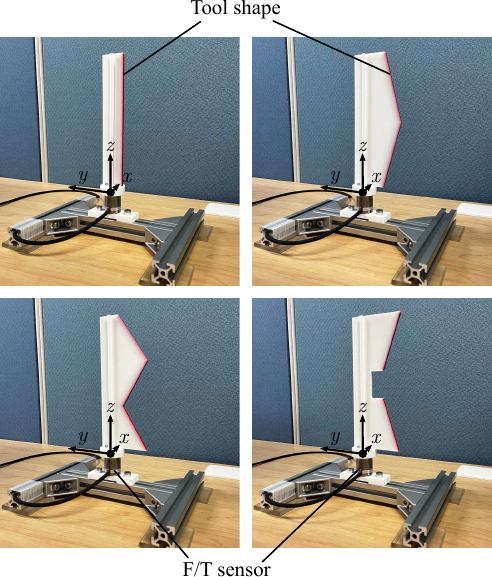}
    \caption{Tools for the experiment.
        \textbf{Upper left:} Straight shape; \textbf{Upper right:} Angular shape; \textbf{Lower left:} Zigzag shape; \textbf{Lower right}: Discontinuous shape. Contact occurs only at the red-painted edges.
        \label{fig:tool for experiment}}
\end{figure}

\begin{table}
    \caption{Specification of the F/T sensor (Source: \url{https://www.leptrino.co.jp/product/6axis-force-sensor}) \label{tab:specification of f/t sensor}}
    \centering
    \begin{tabular}{ll}
        \hline
        Item                    & Value \\
        \hline
        Rated capacity (force)  & $\pm50$ N \\
        Rated capacity (torque) & $\pm0.5$ Nm \\
        Nonlinearity            & $\pm1.0$\% R.O. \\
        Cross-axis sensitivity  & $\pm2.0$\% R.O. \\
        Resolution              & $1 / 4000$ \\
        \hline
    \end{tabular}
\end{table}

\begin{table}
    \caption{Hyper-parameters of the proposed method in the experiments \label{tab:hyper-parameters exp}}
    \small\sf\centering
    \begin{tabular}{lcl}
        \toprule
        Item                       & Symbol                    & Value \\
        \midrule
        Number of particles        & $N$                       & $300$ \\
        Size of grid cells         & -                         & $0.5 \mathrm{~cm}$ \\
        Number of cells            & $N_{\mathrm{s}}$          & $80 \times 80$ \\
        Threshold of resampling    & $N_{\mathrm{th}}$         & $0.207$ \\
        Std. of contact position   & $\sigma_{\bm{c}}$         & $5.64 \times 10^{-4} \mathrm{~cm}$ \\
        Std. of measured moment    & $\sigma_{\bm{M}}$         & $6.90 \times 10^{-5} \mathrm{~Nm}$ \\
        Threshold of distance      & $d_{\mathrm{th}}$         & $0.653 \mathrm{~cm}$ \\
        Threshold of angle         & $\theta_{\mathrm{th}}$    & $0.389 \mathrm{~rad}$ \\
        Increment of shape params. & $\Delta s_{\mathrm{inc}}$ & $0.0609$ \\
        Decrement of shape params. & $\Delta s_{\mathrm{dec}}$ & $0.0285$ \\
        \bottomrule
    \end{tabular}
\end{table}

\begin{table}
    \caption{Hyper-parameters of the baseline method in the experiments, where the symbols follow \cite{Tsuji2017} \label{tab:hyper-parameters baseline exp}}
    \small\sf\centering
    \begin{tabular}{lcl}
        \toprule
        Item                  & Symbol   & Value \\
        \midrule
        Forgetting factor     & $\rho$   & $0.979$ \\
        Regularization factor & $\alpha$ & $1.17 \times 10^{4}$ \\
        \bottomrule
    \end{tabular}
\end{table}

Experiments were conducted using tools attached to a force/torque sensor, as shown in Figure~\ref{fig:tool for experiment}.
Although the sensor is on a fixture instead of a robot, this setup is equivalent to a situation in which a robot with an F/T sensor grips a tool (as illustrated in Figure~\ref{fig:method-overview}) while gravity and inertial forces are eliminated.
It should be possible to extend this setup to dynamic situations by identifying the inertia of the tool in advance using some method, such as \cite{Atkeson1986}, and compensating for the gravity and inertial force.
However, even with this setup, we can evaluate the essential topic of the present issue about how much the proposed method can estimate the tool shape and contact position from force signals.

We used a six-axis F/T sensor (PFS020YA500U6) supplied by Leptrino, Inc.
The basic specifications of the sensor are listed in Table~\ref{tab:specification of f/t sensor}.
In the experiments, we only considered the $y$--$z$ plane.
The tools were straight- and angular-shaped with 15~cm long.

During measurement, the experimenter touched arbitrary positions on the edge of the tool and applied a contact force in various directions.
Contact occurs at a single contact point at a time without large spin torque to maintain (\ref{eqn:equilibrium of moment with measured force}).
Similar to the simulation, the contact force fluctuated for approximately 20~s and after that stopped fluctuating as much as possible.

\begin{figure*}
    \centering
    \includegraphics[scale=0.95]{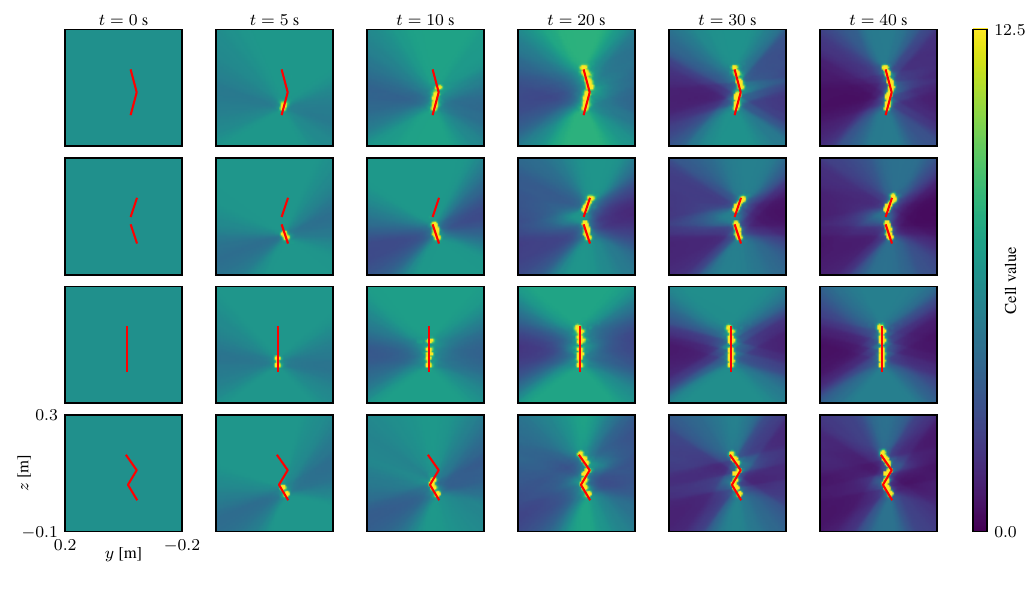}
    \caption{Progress of shape parameter estimation in the experiment.
        Red curves indicate the true object shape.
        The color maps represent the estimated tool shape, where yellow indicates a high degree of certainty that the tool surface is there (i.e., contact can occur). \label{fig:shape parameter estimation progress exp}}
\end{figure*}

\begin{table}
    \caption{Shape-estimation errors in the experiments (means and standard deviations of 5 trials of estimation with the same data) \label{tab:shape-estimation errors exp}}
    \small\sf\centering
    \begin{tabular}{lcc}
        \toprule
        Shape         & Errors [cm] \\
        \midrule
        Straight      & $ 0.395 \pm 0.092$ \\
        Angular       & $ 0.422 \pm 0.116$ \\
        Zigzag        & $ 0.701 \pm 0.080$ \\
        Discontinuous & $ 0.476 \pm 0.135$ \\
                \bottomrule
    \end{tabular}
\end{table}

\begin{figure*}
    \centering
    \subfloat[Proposed method]{
        \includegraphics{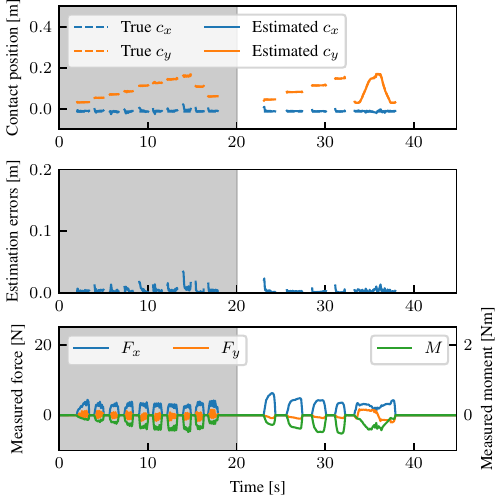}
    }
    \subfloat[Baseline method]{
        \includegraphics{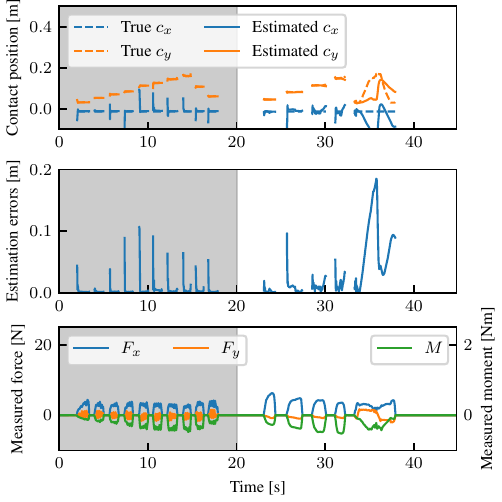}
    }
    \caption{Estimated contact positions in the experiments with a straight-shaped tool.
        Dashed curves are estimated by the method with the tool shape being known \citep{Salisbury1984}.
        The contact force intentionally fluctuated before 20~s (gray area), and then was maintained constant. \label{fig:estimated position results exp straight}}
\end{figure*}

\begin{figure*}
    \centering
    \subfloat[Proposed method]{
        \includegraphics{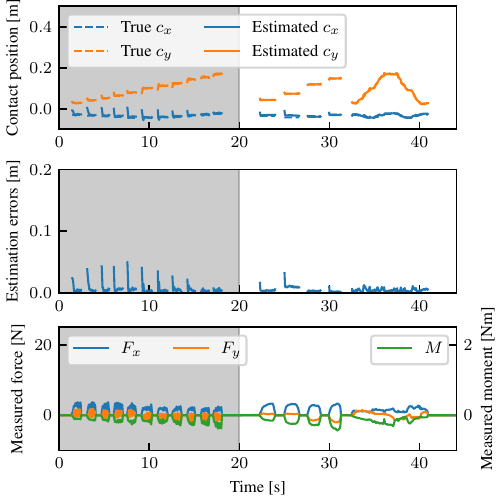}
    }
    \subfloat[Baseline method]{
        \includegraphics{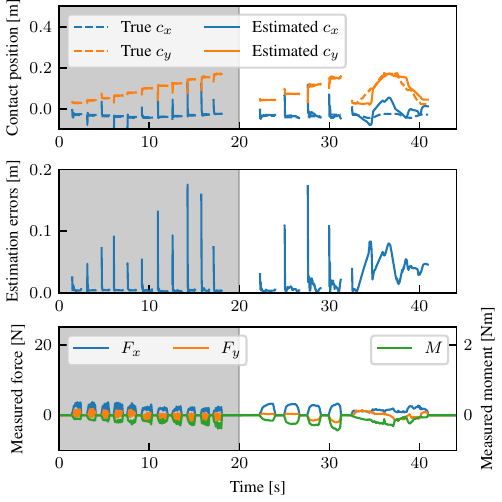}
    }
    \caption{Estimated contact positions in the experiments with an angular-shaped tool.
        Dashed curves are estimated by the method with the tool shape being known \citep{Salisbury1984}.
        The contact force intentionally fluctuated before 20~s (gray area), and then was maintained constant. \label{fig:estimated position results exp angular}}
\end{figure*}

\begin{figure*}
    \centering
    \subfloat[Proposed method]{
        \includegraphics{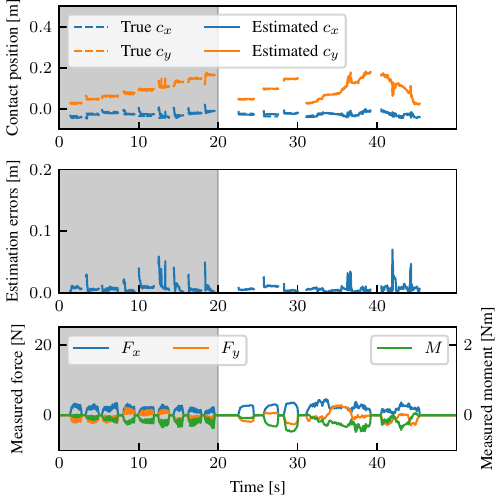}
    }
    \subfloat[Baseline method]{
        \includegraphics{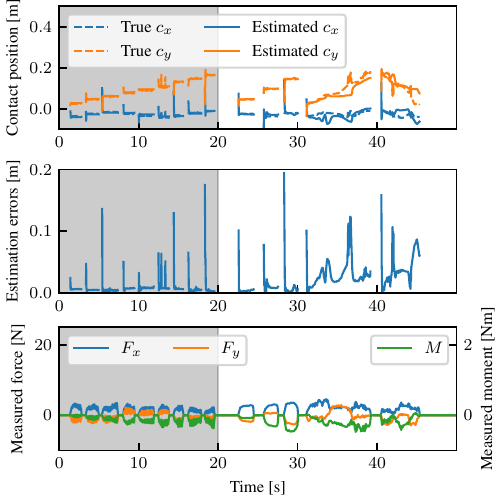}
    }
    \caption{Estimated contact positions in the experiments with a zigzag-shaped tool.
        Dashed curves are estimated by the method with the tool shape being known \citep{Salisbury1984}.
        The contact force intentionally fluctuated before 20~s (gray area), and then was maintained constant. \label{fig:estimated position results exp zigzag}}
\end{figure*}

\begin{figure*}
    \centering
    \subfloat[Proposed method]{
        \includegraphics{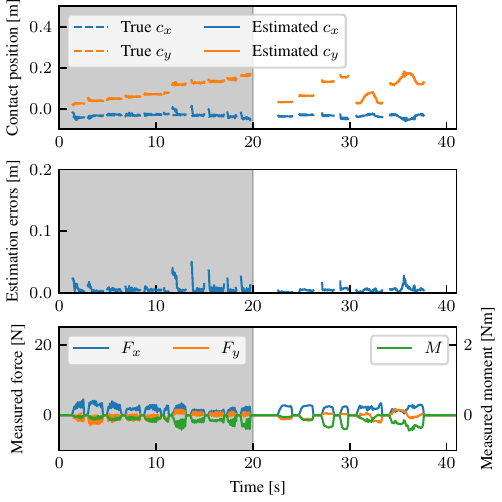}
    }
    \subfloat[Baseline method]{
        \includegraphics{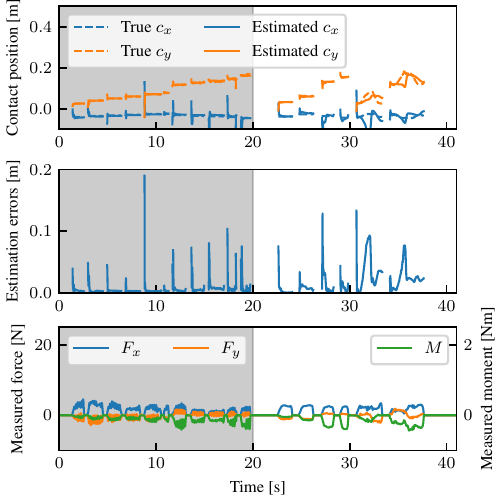}
    }
    \caption{Estimated contact positions in the experiments with a discontinuous-shaped tool.
        Dashed curves are estimated by the method with the tool shape being known \citep{Salisbury1984}.
        The contact force intentionally fluctuated before 20~s (gray area), and then was maintained constant. \label{fig:estimated position results exp discontinuous}}
\end{figure*}

In estimation, we regarded contact occurs when $\|\bm{F}\| \geq 0.5 \mathrm{~N}$.
Once contact is lost, the estimation pauses, and the estimator is initialized except for the shape parameters.
Hyper-parameters are listed in Tables~\ref{tab:hyper-parameters exp} and \ref{tab:hyper-parameters baseline exp}, that were selected by using Optuna \citep{optuna}.

\subsection{Results}

Figure~\ref{fig:shape parameter estimation progress exp} shows the progress of the shape parameter estimation.
The snapshots demonstrate that the tool-shape parameters gradually converged to near the true value.
Table~\ref{tab:shape-estimation errors exp} shows shape-estimation errors in the proposed method in each shape.
The estimated shapes were generally within 1~cm of the actual shapes.

Figures~\ref{fig:estimated position results exp straight}--\ref{fig:estimated position results exp discontinuous} show the results of the contact-position estimation.
Here, we used the oracle method for computing the actual contact positions.
As a result, the proposed method could estimate contact positions with smaller errors than the baseline method.
The contact position can be estimated even when the contact position moves rapidly without contact force fluctuations (after approximately 30~s).
This is difficult for the baseline method because it requires contact force fluctuation and slow contact-position movement.

\section{Discussion}

The proposed method can estimate the contact position and tool shape starting from unknown values.
The estimation process is as follows:
The proposed method first estimates the contact position using the principle in \cite{Tsuji2017}.
This strategy is based on the assumption that the contact position moves slowly as contact force fluctuates.
Using the estimated contact positions, the proposed method gradually estimated the tool shape ($t < 10 \mathrm{~s}$ in Figure~\ref{fig:shape parameter estimation progress}).
Once tool shape estimation converged ($t \geq 10 \mathrm{~s}$ in Figure~\ref{fig:shape parameter estimation progress}), the proposed method can estimate contact position without contact force fluctuation.
In addition, the proposed method does not need to switch between tool-shape estimation and contact-position estimation explicitly because both are performed simultaneously based on a framework of probabilistic inference.
Although the estimation tends to proceed sequentially as mentioned above, both estimation processes always run simultaneously, and there is no need for those two steps to be clearly separated.
Thus the proposed method is applicable to the situations where the contact force does not always fluctuate during tasks.
This contrasts with the baseline method (Figure~\ref{fig:estimated position results}\subref{fig:estimated position results baseline}), in which the contact-position estimation largely oscillated after the contact force fluctuation stopped.
The same tendency was observed in the experimental results with an actual F/T sensor (Figures~\ref{fig:shape parameter estimation progress exp}--\ref{fig:estimated position results exp discontinuous}).

The proposed method can estimate the tool shape regardless of the parameter dimensions (i.e., grid resolutions), as shown in Figure~\ref{fig:results with varying resolutions}.
It is worth noting that the number of particles can be far smaller than the parameter dimensions.
It was possible to handle shape parameters of the order of $10^5$ by using the particle number of the order of only $10^2$.
Although increased shape parameters needs a large computational cost, it will be still far efficient than the naive particle filtering and could be made more efficient by GPU parallelization.
By contrast, some other hyper-parameters affect the estimation performance.
First, the number of particles, $N$, strongly affects the performance, according to Figure~\ref{fig:position errors results with varying particles}.
With a smaller number of particles it is easier to fall into a sub-optimal local minimum, as there may be insufficient particle around appropriate estimates during the initial stages of estimation.
Second, too small $d_\mathrm{th}$ and $\Delta s_{\mathrm{inc}}$ and too large $\theta_\mathrm{th}$ may often result in collapsed shape estimation, according to Figures~\ref{fig:estimated shape parameters with varying params} and \ref{fig:estimated shape parameters with varying delta s}.
Also, a large $\theta_\mathrm{th}$ is considered to adversely affect the estimation for tool shapes with high curvature by decreasing cell values over a wide range.
A small $\theta_\mathrm{th}$, on the other hand, slightly delayed the convergence of estimation.
Third, preferred $\Delta s_{\mathrm{dec}}$ seem to depend on the magnitude of $\Delta s_{\mathrm{inc}}$, according to Figure~\ref{fig:estimated shape parameters with varying delta s}.

Comparison of the proposed method with the naive method suggests the effectiveness of the proposed shape-parameter update method, even apart from the dimensionality issue described above.
The naive method resulted in poor performance in shape estimation even in a low-dimensional case, as shown in Figure~\ref{fig:shape parameter estimation progress naive}.
This would be because the naive particle filter treats variables (i.e. cells of the shape parameter and the contact position coordinates) as independent each other, ignoring their geometric relationship.
Thus, the naive particle filter need to explore the entire space of the shape parameters including cells in a position that has nothing to do with the current contact positions.
This is inefficient and would be easy to converge to a bad local minima.
This may highlight the effectiveness of the proposed shape update method ((\ref{eqn:shape estimation criterion 1}), (\ref{eqn:shape estimation criterion 2}), and Figure~\ref{fig:shape computation}) that utilize the intrinsic contact sensing principle formulated in (\ref{eqn:line of action}).

There are some limitations to this study.
First, the proposed method still requires fluctuations in the contact force at the beginning of estimation to estimate the tool shape accurately.
As Figure~\ref{fig:position errors results with varying fluctuations} shows, less fluctuation before shape estimation converges results in worse contact-position estimation.
This would be a theoretical limitation of methods using force signal-based inference.
Second, the proposed method must be extended to tools with more complex shapes.
This requires a more efficient estimation and modified shape-estimation criteria.
For instance, although the shape estimation criterion defined in (\ref{eqn:shape estimation criterion 2}) accelerates shape estimation, it assumes that there is only a single intersection point between the tool surface and the line of action defined in (\ref{eqn:line of action}), making it inherently impossible to apply it to non-convex shapes.
For practical applications, exception handling is also desired to reduce the impact of sudden changes in contact position or multiple points of contact on shape estimation, that affect adversely.
Even with these weaknesses, however, this study demonstrated that the proposed method can estimate the tool shape with a large number of parameters only from force signals.
We believe that this study has shown new development and challenges in force-signal processing.

We also have future works for a combination with robot controls.
Improvement of accuracy would be needed for application to daily tasks, which sometimes require sub-centimeter accuracy.
This could be achieved by a further research of the shape update rule.
Force control to realize stable contact with the environment in the initial stages of estimation for estimation should be researched.
Also, a control strategy that can reduce uncertain regions efficiently would make shape estimation faster than random contact as performed in this study.
As the proposed method is based on probabilistic inference, it can be integrated with exploration strategies into a single probabilistic inference.
This idea associates the proposed method with \emph{control as inference} \citep{Levine2018} and \emph{active inference} \citep{Friston2016}, which have garnered attention in the field of robotics.
Besides, use of prior knowledge of other modalities, such as vision, would be desired to realize more rapid and stable estimation.

\section{Conclusion}

In this study, we propose a method for simultaneously estimating the contact position and tool shape using a grid representation.
Using the proposed probabilistic model and shape parameter update method, the proposed method can estimate the shape representation with even more than $10^5$ dimensions using only 300 particles.
The proposed method was evaluated using simulations and experiments.
Consequently, it can estimate the tool-shape geometry simultaneously with the contact positions.
Moreover, owing to shape estimation, contact positions can be accurately estimated when the contact force does not fluctuate, which is impossible to achieve using the conventional methods without shape information or shape estimation.

\begin{dci}
The author(s) declared no potential conflicts of interest with respect to the research, authorship, and/or publication of this article.
\end{dci}

\begin{funding}
This work was supported by JSPS KAKENHI Grant Number 22K14212.
\end{funding}

\begin{sm}
Supplemental material for this article is available online.
The code is readily available at \url{https://doi.org/10.5281/zenodo.16949213}.
\end{sm}

\end{document}